\documentclass{article}
\usepackage{iclr2026_conference,times}

\usepackage{amsmath,amsfonts,bm}

\def\eqref#1{equation~\ref{#1}}

\def\1{\bm{1}}

\DeclareMathAlphabet{\mathsfit}{\encodingdefault}{\sfdefault}{m}{sl}
\SetMathAlphabet{\mathsfit}{bold}{\encodingdefault}{\sfdefault}{bx}{n}

\usepackage{hyperref}
\usepackage{url}
\usepackage{booktabs}
\usepackage{etoolbox}
\usepackage{amsfonts}
\usepackage{nicefrac}
\usepackage{microtype}
\usepackage{xcolor}
\usepackage{colortbl}
\usepackage{amsmath}
\usepackage{graphicx}
\usepackage{wrapfig}
\usepackage{fvextra}
\usepackage{enumitem}
\usepackage{caption}
\usepackage[most]{tcolorbox}
\usepackage{tabularx}
\usepackage{amsthm}
\usepackage{bm}
\usepackage{multirow}
\tcbuselibrary{breakable}
\usepackage[capitalize,noabbrev]{cleveref}
\usepackage{titletoc}
\usepackage{tikz}

\definecolor{mintblue}{RGB}{210,235,250}
\definecolor{mintframe}{RGB}{120,180,220} 
\definecolor{minttitle}{RGB}{100,150,200} 
\definecolor{minttext}{RGB}{50,80,120}    

\definecolor{runzhemilk}{RGB}{255,235,245} 
\definecolor{roseframe}{RGB}{230,120,150}  
\definecolor{runzhecotton}{RGB}{255,170,200}    

\newtcolorbox{promptbox}[1]{
  enhanced,
  breakable,
  colback= runzhemilk!30!white,   
  colframe=roseframe,                
  colbacktitle= runzhecotton!66!white, 
  coltitle=white!33,
  title=\textbf{#1},
  fonttitle=\bfseries,
  sharp corners=south, 
  borderline={0.8pt}{0pt}{roseframe},
  boxrule=0.8pt,
  arc=6pt, 
  left=6pt, right=6pt, top=6pt, bottom=6pt,
  before skip=10pt, after skip=10pt,
  drop shadow=black!12,      
}

\newtcolorbox{casebox}[1]{
enhanced,
breakable,
colback=mintblue!40!white,
colframe=mintframe,
colbacktitle=minttitle!70!white,
coltitle=white,
title=\textbf{#1},
fonttitle=\bfseries,
sharp corners=south, 
borderline={0.8pt}{0pt}{minttitle},
boxrule=0.8pt,
arc=6pt, 
left=6pt, right=6pt, top=6pt, bottom=6pt,
before skip=10pt, after skip=10pt,
drop shadow=black!15, 
}

\newtcolorbox{takeawaysbox}{
enhanced,
breakable,
colback=mintblue!40!white,
colframe=mintframe,
colbacktitle=minttitle!70!white,
coltitle=white,
title=\textbf{Key Takeaways},
fonttitle=\bfseries,
sharp corners=south, 
borderline={0.8pt}{0pt}{minttitle},
boxrule=0.8pt,
arc=6pt, 
left=6pt, right=6pt, top=6pt, bottom=6pt,
before skip=10pt, after skip=10pt,
drop shadow=black!15, 
}

\makeatletter
\newcommand{\DrawLine}{%
  \begin{tikzpicture}
  \path[use as bounding box] (0,0) -- (\linewidth,0);
  \draw[color=minttitle!70!white,dashed,dash phase=1.5pt]
        (0-\kvtcb@leftlower-\kvtcb@boxsep,0)--
        (\linewidth+\kvtcb@rightlower+\kvtcb@boxsep,0);
  \end{tikzpicture}%
  }
\makeatother

\definecolor{darkblue}{rgb}{0, 0, 0.5}
\definecolor{citeblue}{rgb}{0.35, 0.62, 0.88}
\hypersetup{colorlinks=true, citecolor=citeblue, urlcolor=citeblue, linkcolor=darkblue}

\newcommand{\projectpageurl}{}
\newcommand{\huggingfaceurl}{https://huggingface.co/collections/bingyang-lei/draft-opd}
\newcommand{\githuburl}{https://github.com/bingyang-lei/Draft-OPD}
\newcommand{\projecturl}{https://www.haodilei.top/draft-opd}
\newcommand{\resourceIcon}[1]{\raisebox{-0.24em}{\includegraphics[height=1.05em]{#1}}}
\newcommand{\resourceLabel}[2]{\resourceIcon{#1}\hspace{0.35em}\textcolor{citeblue}{\textbf{#2}}}

\newcommand{\resourceLink}[3]{\href{#1}{\resourceLabel{#2}{#3}}}

\title{Draft-OPD: On-Policy Distillation for Speculative Draft Models}

\author{%
\begin{minipage}{0.96\textwidth}
\vspace*{0.6em}
\centering
\normalfont
{\small
\textbf{Haodi Lei}\textsuperscript{1,2}\enspace
\textbf{Yafu Li}\textsuperscript{2,4,\textdagger}\enspace
\textbf{Haoran Zhang}\textsuperscript{1,2}\enspace
\textbf{Shunkai Zhang}\textsuperscript{2,5}\enspace
\textbf{Qianjia Cheng}\textsuperscript{2,6}\\
\textbf{Xiaoye Qu}\textsuperscript{2}\enspace
\textbf{Ganqu Cui}\textsuperscript{2}\enspace
\textbf{Bowen Zhou}\textsuperscript{2,3}\enspace
\textbf{Ning Ding}\textsuperscript{3,2,\textdagger}\enspace
\textbf{Yun Luo}\textsuperscript{2,\textdagger}\enspace
\textbf{Yu Cheng}\textsuperscript{4,2}
}\\[0.35em]
{\footnotesize
\textsuperscript{1}\,Shanghai Jiao Tong University \quad
\textsuperscript{2}\,Shanghai AI Laboratory \quad
\textsuperscript{3}\,Tsinghua University\\
\textsuperscript{4}\,The Chinese University of Hong Kong \quad
\textsuperscript{5}\,Peking University \quad
\textsuperscript{6}\,Zhejiang University
}
\end{minipage}%
}

\iclrfinalcopy

\begin{document}

\maketitle

\begingroup
\renewcommand{\thefootnote}{\fnsymbol{footnote}}
\footnotetext[2]{Corresponding authors.}
\endgroup

\begin{abstract}
Speculative decoding accelerates large language model inference by pairing a target model with a lightweight draft model whose proposed tokens are verified in parallel. A common way to build draft models, like EAGLE3 or DFlash is supervised fine-tuning (SFT) on target-generated trajectories. However, we observe that SFT quickly plateaus: the draft model's acceptance length on test data stops improving. The reason is an offline-to-inference mismatch: In SFT, the drafter learns from fixed target-generated trajectories, whereas during speculative decoding it is evaluated on blocks proposed under its own policy. This motivates on-policy distillation (OPD), where the target model supervises the drafter on draft-induced states. Yet OPD remains difficult for draft models, as they cannot reliably roll out complete sequences independently, whereas target-assisted generation makes the collected sequences follow the target distribution and thus eliminates the on-policy signal. We therefore propose Draft-OPD, which uses target-assisted rollout for stable continuations and replays drafting from the verification-exposed error positions. This allows the drafter to learn from target feedback on both accepted and rejected proposals, focusing training on the draft-induced errors that limit speculative acceptance. Experiments show that Draft-OPD achieves over $5\times$ lossless acceleration for thinking models across diverse tasks, improving over EAGLE-3 and DFlash by 23\% and 13\%.
\end{abstract}

\begin{center}
{\normalsize
\resourceLink{\projecturl}{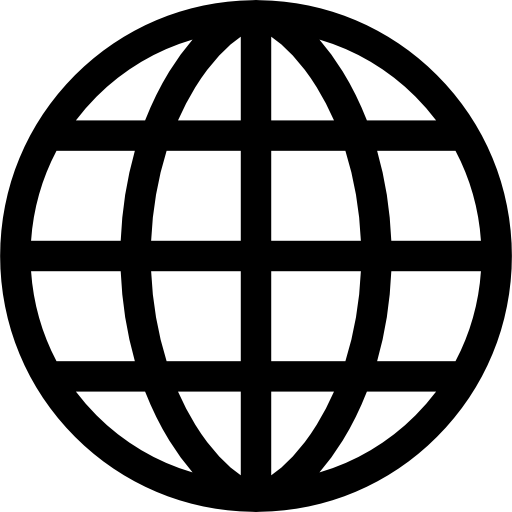}{Project Page}
\hspace{2em}
\resourceLink{\githuburl}{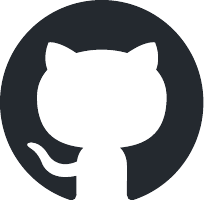}{Code}
\hspace{2em}
\resourceLink{\huggingfaceurl}{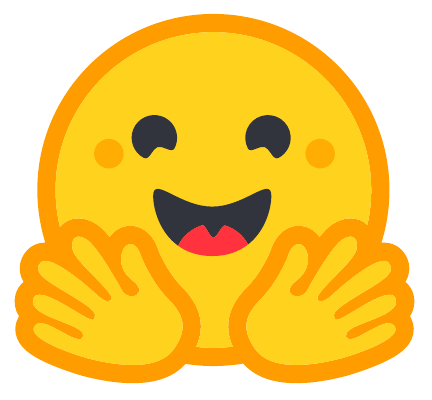}{Models}
}
\end{center}

\section{Introduction}
\label{sec:introduction}

\begin{wrapfigure}{r}{0.5\textwidth}
  \vspace{-2.0em}
  \centering
  \includegraphics[width=\linewidth]{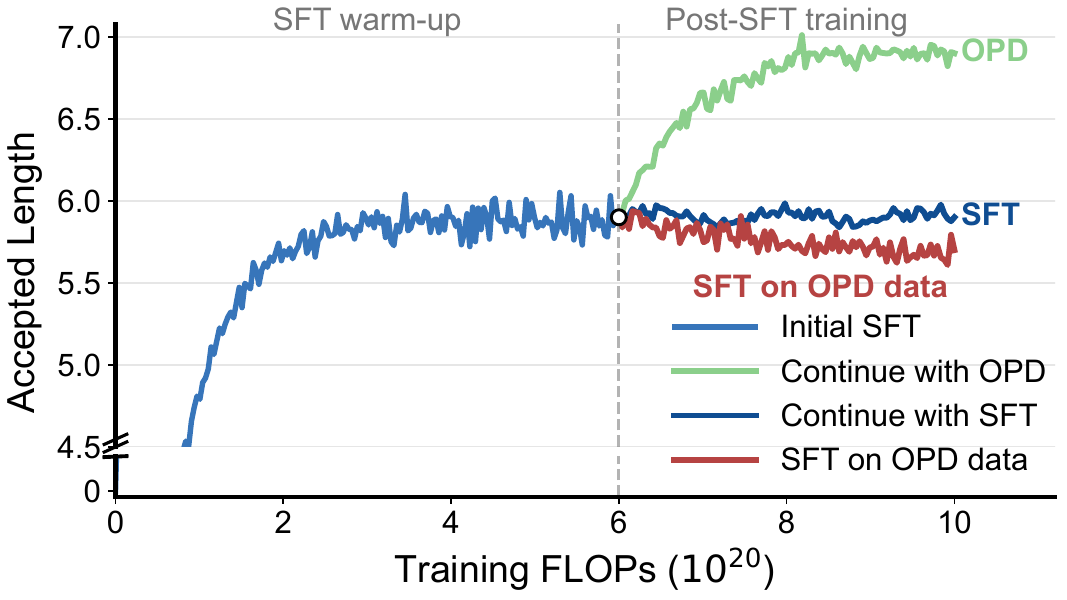}
  \caption{Accepted length during draft-model training. After an initial SFT warm-up, continued offline SFT quickly plateaus, and simply applying OPD data for SFT can even reduce accepted length. In contrast, Draft-OPD continues to improve accepted length by online training.}
  \label{fig:accepted-length}
\end{wrapfigure}

Recent advances in large language models (LLMs) have enabled strong performance across reasoning, coding, and general assistant tasks, but their growing model sizes and longer generations substantially increase inference cost \citep{yang2025qwen3,guo2025deepseekr1}. Speculative decoding (SD) mitigates this cost by using a lightweight draft model to propose tokens that are verified in parallel by a larger target model, preserving the target model's output distribution \citep{leviathan2023speculative,chen2023speculative}. In SD, the speedup depends heavily on the quality of the draft model: when the drafter closely matches the target model, longer token spans can be accepted, reducing the number of expensive target-model decoding steps. Recent mainstream methods, such as EAGLE-3 and DFlash, have achieved strong acceleration by training lightweight draft models on target-generated trajectories.
\citep{li2025eagle3,chen2026dflash}.

However, we find that offline supervised fine-tuning quickly reaches its limit for draft-model training. Figure~\ref{fig:accepted-length} shows a representative training curve. After an initial SFT warm-up, continuing SFT does not improve the draft model's acceptance length on test data; instead, the accepted length fluctuates around a fixed plateau. Moreover, continuing SFT on the data used for OPD can even reduce the accepted length. This suggests that the key limitation does not lie in offline training compute. Instead, the plateau points to a mismatch between the states used for training and the states that determine speculative acceptance. In SFT, the drafter learns from fixed target-generated trajectories, so every prefix is produced by the target model. During speculative decoding, however, the target model verifies token blocks proposed by the drafter itself. The accepted length is therefore determined by draft-induced inference states, rather than only by the static states seen in offline target trajectories.

This offline-to-inference mismatch motivates on-policy distillation (OPD), where the target model supervises states induced by the current draft policy \citep{agarwal2024onpolicy}. Directly applying OPD to draft models, however, is not straightforward. Standard OPD assumes that the student can roll out full sequences under its own policy. But mainstream draft models such as EAGLE-style or DFlash-style drafters are designed to propose short token spans under target-model guidance, rather than to act as standalone autoregressive generators. As a result, draft-only rollouts can easily become repetitive or low quality. Using target-assisted rollouts can produce usable sequences, but the verified continuation follows the target distribution: target model discards the incorrect tokens proposed by draft model, even though these errors are the most valuable training signals. Consequently, draft model still learns on trajectories generated by target model rather than on its own inference-time states.

To address this gap, we propose \textbf{Draft-OPD}, an on-policy distillation framework designed for speculative draft models. Draft-OPD uses target-assisted rollout to maintain stable continuations, and records error positions exposed by speculative verification. It then replays drafting from these positions and asks the target model to score the same draft-generated prefixes. This makes it possible to train on states where the draft model actually acted, including both accepted proposals and rejected proposals. Finally, Draft-OPD uses an acceptance-aware distillation objective that treats these two token groups differently: accepted tokens reinforce reliable agreement with the target model, while rejected tokens focus learning on draft-induced errors that limit speculative acceptance.

In summary, our contributions are:

\begin{itemize}

\item We identify a key limitation of offline SFT for draft models: SFT quickly plateaus because the drafter is trained on fixed target trajectories but evaluated on blocks induced by its own policy during speculative decoding.

\item We explain why standard OPD does not directly apply to draft models: draft-only rollouts are unstable, while target-assisted rollouts remove the on-policy signal.

\item We propose \textbf{Draft-OPD}, an on-policy distillation framework with error-position replay, making it feasible to efficiently post-train draft models on verification-time errors.

\item Experiments show that Draft-OPD achieves over \(5\times\) lossless acceleration for thinking models and improves over EAGLE-3 and DFlash by \(23\%\) and \(13\%\) under matched FLOPs.

\end{itemize}

\section{Related Work}
\label{sec:related-work}

\subsection{Speculative Decoding}
\label{sec:speculative-decoding}

Speculative decoding accelerates autoregressive inference by using a lightweight draft model to propose multiple future tokens and a larger target model to verify them in parallel while preserving the target distribution \citep{leviathan2023speculative,chen2023speculative}. Recent work further improves speculative decoding by leveraging feature-level context from the frozen target model, as in EAGLE \citep{li2024eagle,li2025eagle3}, or by designing stronger block-level draft architectures such as DFlash \citep{chen2026dflash}. These methods substantially improve decoding efficiency, but they still train draft models with offline SFT on target-generated trajectories, which limits further improvements in draft-model acceptance length. In contrast, our work studies how to move draft models beyond this offline training recipe.

\subsection{Draft Model Distillation}
\label{sec:draft-model-distillation}

Because speculative speedup depends on how closely the draft distribution matches the target distribution, several methods train draft models through distillation. DistillSpec aligns a compact draft model with the target model and highlights the importance of on-policy data and task-specific divergence choices \citep{zhou2024distillspec}; temperature-centric distillation studies further show that matching training and decoding configurations can improve speculative decoding under challenging sampling settings \citep{ouyang-etal-2024-temperature}. Online speculative decoding updates draft model from observed query distributions during deployment \citep{liu2023onlineSpeculativeDecoding}, while broader on-policy distillation trains students on self-generated sequences to reduce exposure bias \citep{agarwal2024onpolicy}. However, almost all previous work only studies standalone autoregressive draft models, which are often selected as smaller models from the same family as the target model. Our Draft-OPD is designed for training-based draft models such as DFlash or EAGLE3. These draft models cannot independently generate full student trajectories, making online distillation difficult because training must obtain usable rollouts rather than meaningless repetitions.

\section{Preliminary}
\label{sec:preliminary}

\paragraph{Speculative Decoding.}
Speculative decoding accelerates autoregressive generation by pairing a target model $p_\theta$ with a lightweight draft model $q_\phi$ \citep{leviathan2023speculative,chen2023speculative}. Given a prefix $x_{<t}$, the draft model proposes a block of $K$ candidate tokens,
\begin{equation}
  \begin{aligned}
    \hat{y}_{t+k}
    &\sim
    q_\phi(\cdot \mid x_{<t}, \hat{y}_{t:t+k-1}), \\
    k &= 0,\ldots,K-1.
  \end{aligned}
\end{equation}
and the target model verifies these tokens in parallel. The verifier accepts the longest valid prefix of the drafted block and then continues generation from the verified prefix, preserving the target model's output distribution while reducing the number of expensive target-model decoding steps.

Draft models can be categorized by how they are obtained. Model-based draft models use smaller autoregressive models from the same family as the target model \citep{miao2024specinfer}. A more mainstream approach trains a lightweight draft model specifically for a given target model \citep{cai2024medusa,hui2026peagle,liu2026dart}. These training-based draft models often share the target model's embedding layer and LM head, and are designed to predict short draft spans rather than to generate complete sequences independently. In this work, we focus on training-based draft models because they can provide stronger acceleration.

A central metric for speculative decoding is the accepted length $\tau$, i.e., the number of draft tokens accepted in each verification round. Higher $\tau$ means that each target-model verification generates more tokens, which directly improves decoding efficiency. It also reflects the alignment between the draft and target models, with better alignment yielding longer accepted spans.

\paragraph{On-Policy Distillation.}
On-policy distillation trains a student model on states induced by the student's own policy, rather than only on fixed teacher-generated trajectories \citep{agarwal2024onpolicy}. Given a prompt \(x\), the current student samples a trajectory and induces prefix states
\begin{equation}
  \tilde{y}\sim q_\phi(\cdot\mid x),
  \quad
  s_t=(x,\tilde{y}_{<t}).
\end{equation}
At each state \(s_t\), the teacher provides the next-token distribution. Writing \(p_\theta^t=p_\theta(\cdot\mid s_t)\) and \(q_\phi^t=q_\phi(\cdot\mid s_t)\), a standard OPD objective is
\begin{equation}
  \mathcal{L}_{\mathrm{OPD}}
  =
  \mathbb{E}_{s_t}
  \left[
  D_{\mathrm{KL}}(p_\theta^t\,\|\,q_\phi^t)
  \right].
\end{equation}
Thus, OPD directly targets the exposure mismatch between offline supervised training and inference-time generation: the student learns from states that it actually visits under its own policy.

This perspective is particularly relevant to speculative decoding. As discussed above, the accepted length $\tau$ is determined by how well the draft model aligns with the target model on the prefixes encountered during the draft-verify process. However, SFT trains $q_\phi$ only on fixed target-generated trajectories, whereas speculative decoding evaluates blocks produced by $q_\phi$ itself. OPD offers a natural way to reduce this mismatch by using the target model to supervise draft-induced states, particularly those induced by the draft model’s own deviations from the target.

\section{Method}
\label{sec:method}

\subsection{Challenges of Direct OPD}
\label{sec:direct-opd-challenge}

\begin{wrapfigure}{r}{0.5\textwidth}
  \centering
  \includegraphics[width=\linewidth]{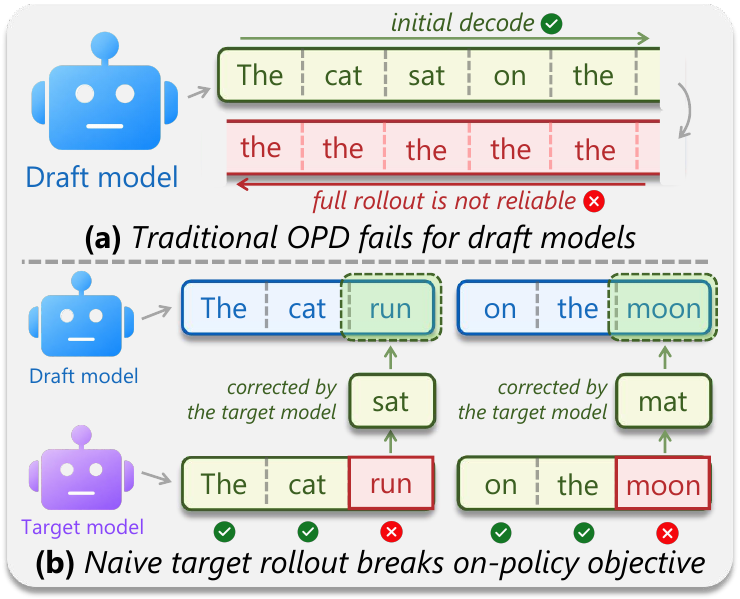}
  \caption{Why direct OPD is unsuitable for draft models. (a) Trajectories from draft model are repetitive. (b) Naive target-assisted rollout makes the sequence follow target distribution, losing informative errors from rejected tokens.}
  \label{fig:draft-opd-failure}
\end{wrapfigure}

Applying OPD to draft models requires both stable rollouts and draft-induced training states. Standard OPD assumes that the student can roll out full sequences under its own policy, but EAGLE- and DFlash-style draft modules are designed to propose short token blocks under target-model verification, rather than to act as standalone autoregressive generators. As shown in Figure~\ref{fig:draft-opd-failure}(a), forcing such draft modules to self-rollout full trajectories can produce repetitive or degenerate samples, making the resulting supervision unreliable.

Target-assisted rollout avoids degenerate samples, but it removes the draft-policy signal that OPD is meant to capture. As illustrated in Figure~\ref{fig:draft-opd-failure}(b), strict speculative verification is lossless with respect to the target model, so the verified continuation follows the target distribution rather than the draft policy. Moreover, this brings back the offline-to-inference mismatch: target-assisted keeps accepted tokens and discards rejected proposals, even though these rejected tokens reveal the draft model's most informative errors. We empirically verify this limitation in Section~\ref{sec:target-assisted-rollout-analysis}, where naive target-assisted rollout underperforms Draft-OPD. A suitable OPD method for draft modules must therefore keep target-quality rollouts while preserving the draft-induced errors exposed during verification.

\subsection{Draft-OPD}
\label{sec:draft-opd}

Motivated by these failure modes, we design Draft-OPD to enable effective on-policy training for draft models. The key idea is to use target assistance to keep rollouts stable, while replaying draft-induced errors to preserve the on-policy training signal. Draft-OPD implements this idea through three coordinated designs: rollout with error-position collection, replay for log-probability computation, and an acceptance-aware distillation objective.

\paragraph{Rollout with error-position collection.} Let $p_\theta$ denote the target model and $q_\phi$ denote a lightweight draft model initialized from supervised draft-model training. Given a prompt $x$, speculative decoding asks $q_\phi$ to propose a block of $K$ tokens and uses $p_\theta$ to verify the block in parallel. Draft-OPD uses this interaction to collect a target-quality rollout while recording each draft block's starting position as an anchor for later error replay. Let the verified rollout be $y=(y_0,\ldots,y_{T-1})$. During rollout step $m$, the current verified prefix ends at position $a_m$, where $a_m=-1$ denotes the beginning of the sequence. The draft model proposes a block
\begin{equation}
  d_m = (d_{m,1},\ldots,d_{m,K}) \sim q_\phi(\cdot \mid x, y_{\le a_m}),
\end{equation}
and the target model verifies the block. We record $a_m$ as an anchor before moving to the next step. If the verifier accepts $r_m$ tokens from the block, the next anchor is $a_m + r_m$. This process is repeated until the rollout reaches the maximum generation length or the target model emits an end-of-sequence token.

\begin{figure*}[t]
  \centering
  \includegraphics[width=\linewidth]{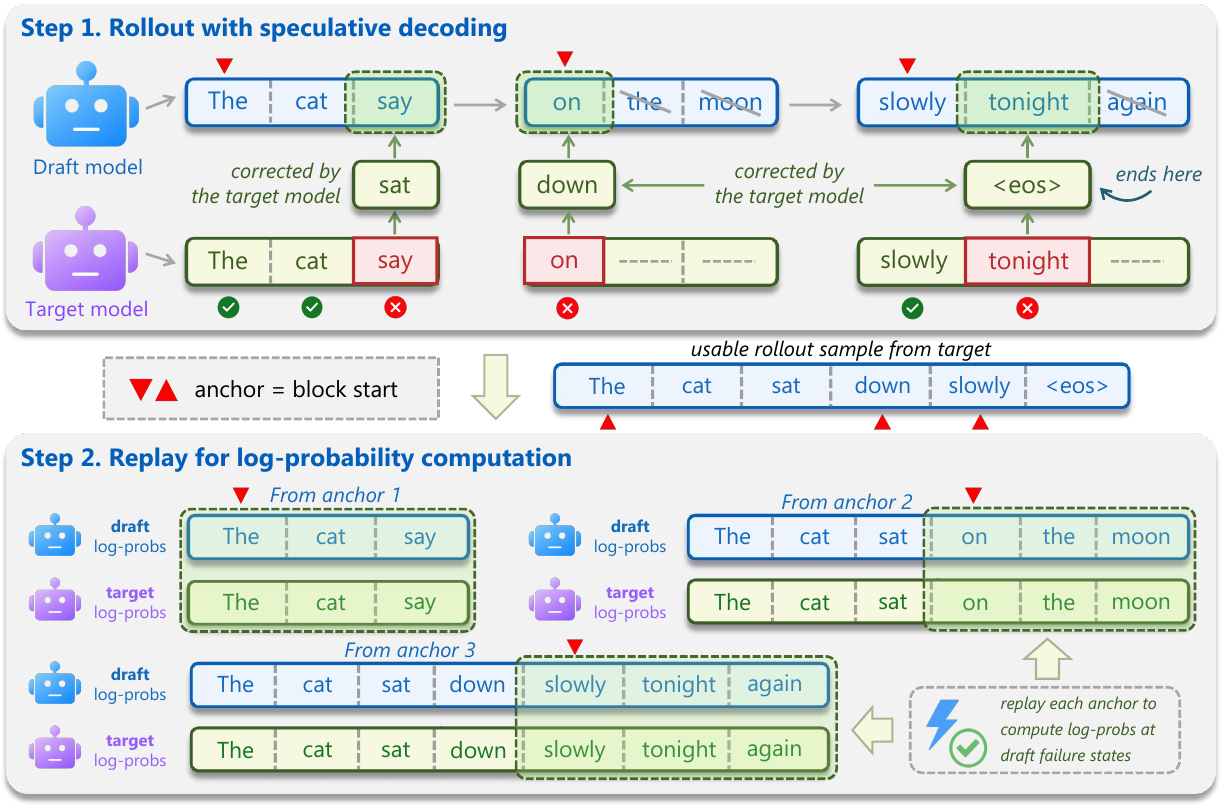}
  \caption{Draft-OPD for draft models. We use speculative decoding to collect stable rollouts and record the start index of each drafted block as an anchor. We then replay drafting from multiple anchors to compute student and teacher log-probabilities for both accepted and rejected draft tokens.}
  \label{fig:draft-opd-method}
\end{figure*}

Anchors preserve the draft model's local actions without requiring the draft model to generate an entire sequence alone. The final rollout remains a high-quality target-model sample, but each anchor identifies a state where the draft model actually proposed a block during inference. Drafting from any anchor requires the target-model hidden states of the preceding tokens; since the rollout has already computed these hidden states on the same sample, Draft-OPD can reuse them during subsequent replay.

\paragraph{Replay for Log-Probability Computation.} After collecting the rollout and anchors, Draft-OPD replays drafting from all anchors to compute the token-level student and teacher log-probabilities. For anchor $a_m$, define the replay context
\begin{equation}
  c_m = (x, y_{\le a_m}).
\end{equation}
Starting from $c_m$, we replay the draft-generated block $d_m$. For each drafted token $d_{m,k}$, the draft model provides the student log-probability, while the target model provides the teacher log-probability on the same draft-generated prefix:
\begin{align}
  \log q_{m,k}(d_{m,k}) &= \log q_\phi(d_{m,k} \mid c_m, d_{m,<k}), \\
  \log p_{m,k}(d_{m,k}) &= \log p_\theta(d_{m,k} \mid c_m, d_{m,<k}).
\end{align}
This replay step differs from training on the final rollout tokens: it evaluates the target model on the draft model's proposed block, including positions that were rejected during verification.

The verification outcome naturally partitions drafted tokens into accepted and rejected sets. If $r_m$ tokens are accepted from block $m$, then
\begin{align}
  \mathcal{I}_{\mathrm{acc}} &= \{(m,k): 1 \le k \le r_m\}, \\
  \mathcal{I}_{\mathrm{rej}} &= \{(m,k): r_m < k \le K\}.
\end{align}
Accepted tokens reflect states where the draft model agrees with the target well enough to pass verification. Rejected tokens include the first failed token and later positions in the same draft block, which become less reliable because an earlier rejection invalidates the remaining drafted suffix.

\paragraph{Acceptance-Aware Distillation Objective.} We use different KL directions for accepted and rejected draft tokens. For accepted tokens, we use forward KL to make the draft distribution cover the target distribution at states where the draft model is already close to the target:
\begin{equation}
  \mathcal{L}_{\mathrm{acc}}
  =
  \frac{1}{|\mathcal{I}_{\mathrm{acc}}|}
  \sum_{(m,k)\in\mathcal{I}_{\mathrm{acc}}}
  D_{\mathrm{KL}}\!\left(p_{m,k}\,\|\,q_{m,k}\right).
\end{equation}
For rejected tokens, we use reverse KL to penalize the draft model's own high-probability modes when the target model disagrees:
\begin{equation}
  \mathcal{L}_{\mathrm{rej}}
  =
  \frac{1}{Z}
  \sum_{(m,k)\in\mathcal{I}_{\mathrm{rej}}}
  w_k
  D_{\mathrm{KL}}\!\left(q_{m,k}\,\|\,p_{m,k}\right),
\end{equation}
where $Z=\sum_{(m,k)\in\mathcal{I}_{\mathrm{rej}}} w_k$ normalizes the rejected-token weights. We give the detailed rationale for this KL design in Appendix~\ref{sec:loss-design}.

Rejected tokens at later positions in a draft block are less important than earlier rejected tokens. In speculative decoding, an early error prevents the verifier from using the remaining suffix, so mistakes near the beginning of a block have a larger effect on acceptance length. We reflect this point with an exponentially decaying weight over block positions:
\begin{equation}
  w_k = \gamma^{k-1}
\end{equation}
The final Draft-OPD objective averages the accepted-token and rejected-token losses:
\begin{equation}
  \mathcal{L}_{\mathrm{Draft\text{-}OPD}}
  =
  \frac{
    \lambda_{\mathrm{acc}}\mathcal{L}_{\mathrm{acc}}
    +
    \lambda_{\mathrm{rej}}\mathcal{L}_{\mathrm{rej}}
  }{
    \lambda_{\mathrm{acc}}+\lambda_{\mathrm{rej}}
  },
  \label{eq:draft-opd-objective}
\end{equation}
We set \(\lambda_{\mathrm{acc}}=\lambda_{\mathrm{rej}}=1\) in all experiments. This objective trains the draft model on stable target-assisted rollouts while preserving the draft-policy errors that determine speculative acceptance.

\section{Experiment}
\label{sec:experiment}
\begin{table*}[t]
  \centering
  \small
  \definecolor{draftopdblue}{RGB}{214,246,248}
  \definecolor{modeband}{RGB}{255,204,204}
  \setlength{\tabcolsep}{4pt}
  \resizebox{\textwidth}{!}{%
  \begin{tabular}{ll*{7}{c}cc}
    \toprule
    \multicolumn{1}{c}{\multirow{3}{*}{\raisebox{-1.0ex}{\textbf{Model}}}}
    & \multicolumn{1}{c}{\multirow{3}{*}{\raisebox{-1.0ex}{\textbf{Method}}}}
    & \multicolumn{7}{c}{\textbf{Speedup}}
    & \multicolumn{2}{c}{\textbf{Mean}} \\
    \cmidrule(lr){3-9}
    \cmidrule(lr){10-11}
    & & \multicolumn{3}{c}{\textbf{\textsc{Math}}}
    & \multicolumn{3}{c}{\textbf{\textsc{Code}}}
    & \multicolumn{1}{c}{\textbf{\textsc{Chat}}}
    & \multirow{2}{*}{\raisebox{-0.45ex}{\textbf{Speedup}}}
    & \multirow{2}{*}{\raisebox{-0.45ex}{\(\boldsymbol{\tau}\)}} \\
    \cmidrule(lr){3-9}
    & & \textbf{GSM8K}
    & \textbf{MATH-500}
    & \textbf{AIME25}
    & \textbf{MBPP}
    & \textbf{HumanEval}
    & \textbf{SWE-Lite}
    & \textbf{MT-Bench}
    & & \\
    \midrule
    \rowcolor{orange!4}
    \multicolumn{11}{c}{Thinking Mode Enabled} \\
    \midrule
    \multicolumn{11}{c}{Temperature $=0$} \\
    \midrule
    \multirow{3}{*}{Q3-4B} & EAGLE-3
    & 4.41$\times$ & 4.15$\times$ & 3.30$\times$ & 4.29$\times$ & 4.38$\times$ & 3.47$\times$ & 3.07$\times$ & 3.87$\times$ & 5.33 \\
    & DFlash
    & 4.51$\times$ & 4.88$\times$ & 4.69$\times$ & 4.39$\times$ & 4.77$\times$ & 4.12$\times$ & 2.96$\times$ & 4.33$\times$ & 5.51 \\
    \rowcolor{draftopdblue}
    & \textbf{Draft-OPD}
    & \textbf{5.31$\times$} & \textbf{5.55$\times$} & \textbf{5.28$\times$} & \textbf{4.85$\times$} & \textbf{5.17$\times$} & \textbf{4.66$\times$} & \textbf{3.18$\times$} & \textbf{4.86$\times$} & \textbf{5.96} \\
    \addlinespace
    \multirow{3}{*}{Q3-8B} & EAGLE-3
    & 4.58$\times$ & 4.45$\times$ & 4.01$\times$ & 4.55$\times$ & 4.46$\times$ & 3.38$\times$ & 3.02$\times$ & 4.06$\times$ & 5.64 \\
    & DFlash
    & 4.67$\times$ & 5.01$\times$ & 4.77$\times$ & 4.50$\times$ & 4.71$\times$ & 3.72$\times$ & 3.01$\times$ & 4.34$\times$ & 5.19 \\
    \rowcolor{draftopdblue}
    & \textbf{Draft-OPD}
    & \textbf{5.36$\times$} & \textbf{5.80$\times$} & \textbf{5.51$\times$} & \textbf{4.86$\times$} & \textbf{5.19$\times$} & \textbf{4.31$\times$} & \textbf{3.18$\times$} & \textbf{4.89$\times$} & \textbf{5.73} \\
    \midrule
    \multicolumn{11}{c}{Temperature $=0.6$} \\
    \midrule
    \multirow{3}{*}{Q3-4B} & EAGLE-3
    & 3.90$\times$ & 3.77$\times$ & 3.07$\times$ & 3.75$\times$ & 3.77$\times$ & 2.75$\times$ & 2.77$\times$ & 3.40$\times$ & 5.03 \\
    & DFlash
    & 4.21$\times$ & 4.46$\times$ & 4.25$\times$ & 3.92$\times$ & 4.17$\times$ & 3.17$\times$ & 2.73$\times$ & 3.84$\times$ & 4.77 \\
    \rowcolor{draftopdblue}
    & \textbf{Draft-OPD}
    & \textbf{4.73$\times$} & \textbf{4.91$\times$} & \textbf{4.56$\times$} & \textbf{4.21$\times$} & \textbf{4.40$\times$} & \textbf{3.38$\times$} & \textbf{2.87$\times$} & \textbf{4.15$\times$} & \textbf{5.13} \\
    \addlinespace
    \multirow{3}{*}{Q3-8B} & EAGLE-3
    & 4.13$\times$ & 4.09$\times$ & 3.57$\times$ & 3.93$\times$ & 3.95$\times$ & 2.83$\times$ & 2.83$\times$ & 3.62$\times$ & \textbf{5.33} \\
    & DFlash
    & 4.23$\times$ & 4.57$\times$ & 4.34$\times$ & 3.85$\times$ & 4.12$\times$ & 2.88$\times$ & 2.76$\times$ & 3.82$\times$ & 4.66 \\
    \rowcolor{draftopdblue}
    & \textbf{Draft-OPD}
    & \textbf{4.77$\times$} & \textbf{5.07$\times$} & \textbf{4.80$\times$} & \textbf{4.17$\times$} & \textbf{4.47$\times$} & \textbf{3.13$\times$} & \textbf{2.91$\times$} & \textbf{4.19$\times$} & 5.05 \\
    \midrule
    \rowcolor{orange!4}
    \multicolumn{11}{c}{Thinking Mode Disabled} \\
    \midrule
    \multicolumn{11}{c}{Temperature $=0$} \\
    \midrule
    \multirow{3}{*}{Q3-4B} & EAGLE-3
    & 4.58$\times$ & 5.70$\times$ & 5.39$\times$ & 4.55$\times$ & 4.53$\times$ & 2.54$\times$ & 2.80$\times$ & 4.30$\times$ & 5.84 \\
    & DFlash
    & 5.36$\times$ & 6.35$\times$ & 5.93$\times$ & 5.00$\times$ & 5.19$\times$ & 3.05$\times$ & 3.01$\times$ & 4.84$\times$ & 6.04 \\
    \rowcolor{draftopdblue}
    & \textbf{Draft-OPD}
    & \textbf{6.22$\times$} & \textbf{7.22$\times$} & \textbf{6.39$\times$} & \textbf{5.40$\times$} & \textbf{5.65$\times$} & \textbf{3.18$\times$} & \textbf{3.09$\times$} & \textbf{5.31$\times$} & \textbf{6.60} \\
    \addlinespace
    \multirow{3}{*}{Q3-8B} & EAGLE-3
    & 4.99$\times$ & 6.07$\times$ & 5.87$\times$ & 4.83$\times$ & 4.94$\times$ & 2.67$\times$ & 3.06$\times$ & 4.63$\times$ & 5.99 \\
    & DFlash
    & 5.69$\times$ & 6.81$\times$ & 6.40$\times$ & 5.17$\times$ & 5.64$\times$ & 3.17$\times$ & 2.92$\times$ & 5.11$\times$ & 6.04 \\
    \rowcolor{draftopdblue}
    & \textbf{Draft-OPD}
    & \textbf{6.49$\times$} & \textbf{7.64$\times$} & \textbf{6.99$\times$} & \textbf{5.64$\times$} & \textbf{6.02$\times$} & \textbf{3.33$\times$} & \textbf{3.12$\times$} & \textbf{5.60$\times$} & \textbf{6.57} \\
    \midrule
    \multicolumn{11}{c}{Temperature $=0.6$} \\
    \midrule
    \multirow{3}{*}{Q3-4B} & EAGLE-3
    & 4.16$\times$ & 4.93$\times$ & 4.49$\times$ & 3.98$\times$ & 4.02$\times$ & 2.41$\times$ & 2.50$\times$ & 3.78$\times$ & 5.66 \\
    & DFlash
    & 5.04$\times$ & 5.87$\times$ & 4.99$\times$ & 4.72$\times$ & 5.17$\times$ & 2.77$\times$ & 2.80$\times$ & 4.48$\times$ & 5.73 \\
    \rowcolor{draftopdblue}
    & \textbf{Draft-OPD}
    & \textbf{5.84$\times$} & \textbf{6.35$\times$} & \textbf{5.31$\times$} & \textbf{4.98$\times$} & \textbf{5.42$\times$} & \textbf{2.90$\times$} & \textbf{2.86$\times$} & \textbf{4.81$\times$} & \textbf{6.13} \\
    \addlinespace
    \multirow{3}{*}{Q3-8B} & EAGLE-3
    & 4.61$\times$ & 5.17$\times$ & 4.85$\times$ & 4.40$\times$ & 4.29$\times$ & 2.43$\times$ & 2.70$\times$ & 4.06$\times$ & 5.69 \\
    & DFlash
    & 5.26$\times$ & 6.18$\times$ & 5.19$\times$ & 4.74$\times$ & 4.97$\times$ & 2.85$\times$ & 2.81$\times$ & 4.57$\times$ & 5.62 \\
    \rowcolor{draftopdblue}
    & \textbf{Draft-OPD}
    & \textbf{6.01$\times$} & \textbf{6.61$\times$} & \textbf{5.57$\times$} & \textbf{5.17$\times$} & \textbf{5.30$\times$} & \textbf{3.03$\times$} & \textbf{2.97$\times$} & \textbf{4.95$\times$} & \textbf{6.02} \\
    \bottomrule
  \end{tabular}%
  }
  \caption{Decoding speedup ratio and average acceptance length ($\tau$) on Qwen3 models. For thinking mode, we use a maximum of 8192 generated tokens; for non-thinking mode, we use a maximum of 2048 generated tokens.}
  \label{tab:main-results}
\end{table*}

\paragraph{Models and tasks.}
We conduct experiments on the Qwen3 family \citep{yang2025qwen3}, including Qwen3-4B, Qwen3-8B, and Qwen3-30B-A3B-Thinking-2507. We evaluated three categories of benchmarks: mathematical reasoning, including GSM8K \citep{cobbe2021gsm8k}, MATH-500 \citep{hendrycks2021math,lightman2023verify} and AIME \citep{aime2025}; code generation and software engineering, including MBPP \citep{austin2021mbpp}, HumanEval \citep{chen2021humaneval} and SWE-bench Lite \citep{jimenez2023swebench}; and out-of-domain benchmark MT-Bench \citep{zheng2023mtbench}.

\paragraph{Datasets.}
For the SFT stage, we use the same training data mixture as DFlash \citep{chen2026dflash}. For the OPD stage, we construct a 16K sample prompt pool randomly sampling 2K prompts from the GSM8K training set, 5K prompts from the MATH corpus after excluding MATH-500 held-out examples, 4K prompts from AoPS \citep{aops}, and 5K prompts from CodeAlpaca \citep{codealpaca}. We use only the questions or instructions from these datasets; responses are generated online by the target model during OPD rather than taken from static reference answers.

\paragraph{Implementation.}
Unless otherwise specified, all experiments are conducted on NVIDIA H200 GPUs, with a batch size of 1, thinking mode enabled. We perform Draft-OPD on top of SFT-trained DFlash draft models: the draft models use 5 Transformer layers for Qwen3-4B and Qwen3-8B, and 8 layers for Qwen3-30B-A3B-Thinking \citep{chen2026dflash}. We use a block size of 16 for both training and inference. Detailed hyperparameters for OPD training are provided in Appendix~\ref{sec:opd-training-details}.

\paragraph{Baselines.}
We compare against EAGLE-3 \citep{li2025eagle3} and DFlash \citep{chen2026dflash}. For a fair comparison, both EAGLE-3 and DFlash are trained with the data mixture introduced by DFlash, and their SFT training budget is matched to the total SFT plus OPD budget of our method so that all draft models are trained under approximately the same FLOPs budget.

\paragraph{Metrics.}
We focus on efficiency-related metrics and do not report generation quality, since Draft-OPD only post-trains the draft model and does not change the speculative decoding procedure used at inference time, thereby preserving the exact output distribution of the target model.
\begin{itemize}
  \item \textbf{Speedup Ratio.} The actual test speedup ratio relative to vanilla autoregressive decoding.
  \item \textbf{Average acceptance length ($\tau$).} The average number of draft tokens accepted by the target model in each verification cycle.
\end{itemize}

\subsection{Main Results}
\label{sec:main-results}

Table~\ref{tab:main-results} compares Draft-OPD with EAGLE-3 and DFlash on Qwen3 models under a matched training FLOPs budget. For EAGLE-3, we use a tree size of 16, with draft steps and top-$k$ set to 8 and 4, respectively. We evaluate both greedy decoding and the recommended Qwen reasoning sampling setting, with temperature 0.6, top-$p$ 0.95, and top-$k$ 20.


\paragraph{Thinking Mode Enabled.} With thinking mode enabled, Draft-OPD consistently improves acceptance length and decoding speed under matched training FLOPs. At temperature 0, it raises the two-model average $\tau$ from 5.35 for DFlash to 5.85 and achieves 4.88$\times$ average speedup across the seven benchmarks, improving over EAGLE-3 and DFlash by 23\% and 13\%.

At temperature 0.6, it remains the fastest method, averaging 4.17$\times$ speedup. Although EAGLE-3 obtains the highest $\tau$ on Qwen3-8B at temperature 0.6, its sequential drafting limits wall-clock speed, supporting our choice to apply OPD to DFlash-style parallel drafting.

\paragraph{Thinking Mode Disabled.} With thinking mode disabled, Draft-OPD preserves the same advantage across decoding temperatures and model sizes. It maintains an average acceptance length of 6.33 and achieves a 5.17$\times$ average speedup. Together with the thinking-mode results, this shows that Draft-OPD improves draft-target alignment across both long reasoning traces and shorter non-thinking generations.

\paragraph{Performance on SGLang.}

\begin{wraptable}{r}{0.5\textwidth}
  \centering
  \small
  \definecolor{tablegreen}{RGB}{30,130,30}
  \setlength{\tabcolsep}{3pt}
  \resizebox{\linewidth}{!}{%
  \begin{tabular}{cc*{5}{c}c}
    \toprule
    \multicolumn{1}{c}{\multirow{2}{*}{\raisebox{-0.45ex}{\textbf{Task}}}}
    & \multicolumn{1}{c}{\multirow{2}{*}{\raisebox{-0.45ex}{\textbf{Method}}}}
    & \multicolumn{5}{c}{\textbf{Concurrency}}
    & \multicolumn{1}{c}{\multirow{2}{*}{\raisebox{-0.45ex}{\textbf{Avg. $\boldsymbol{\tau}$}}}} \\
    \cmidrule(lr){3-7}
    & & \textbf{1} & \textbf{4} & \textbf{8} & \textbf{16} & \textbf{32} & \\
    \midrule
    \multicolumn{8}{l}{\textbf{Qwen3-4B (Enable Thinking)}} \\
    \midrule
    \multirow{3}{*}{AIME25}
    & DFlash & 912 & 2755 & 4750 & 6841 & 8410 & 6.06 \\
    \cmidrule(lr){2-8}
    & \multirow{2}{*}{Draft-OPD}
    & \textbf{969} & \textbf{2984} & \textbf{5071} & \textbf{7297} & \textbf{9043}
    & \multirow{2}{*}{\textbf{6.59}} \\
    & & \textcolor{tablegreen}{(6\%$\uparrow$)} & \textcolor{tablegreen}{(8\%$\uparrow$)} & \textcolor{tablegreen}{(7\%$\uparrow$)} & \textcolor{tablegreen}{(7\%$\uparrow$)} & \textcolor{tablegreen}{(8\%$\uparrow$)} & \\
    \midrule
    \multirow{3}{*}{MATH-500}
    & DFlash & 949 & 3157 & 5234 & 8004 & 10062 & 6.17 \\
    \cmidrule(lr){2-8}
    & \multirow{2}{*}{Draft-OPD}
    & \textbf{1036} & \textbf{3413} & \textbf{5603} & \textbf{8593} & \textbf{10943}
    & \multirow{2}{*}{\textbf{6.68}} \\
    & & \textcolor{tablegreen}{(9\%$\uparrow$)} & \textcolor{tablegreen}{(8\%$\uparrow$)} & \textcolor{tablegreen}{(7\%$\uparrow$)} & \textcolor{tablegreen}{(7\%$\uparrow$)} & \textcolor{tablegreen}{(9\%$\uparrow$)} & \\
    \midrule
    \multirow{3}{*}{SWE-Lite}
    & DFlash & 901 & 2983 & 5009 & 7743 & 9604 & 5.62 \\
    \cmidrule(lr){2-8}
    & \multirow{2}{*}{Draft-OPD}
    & \textbf{976} & \textbf{3230} & \textbf{5544} & \textbf{8568} & \textbf{10538}
    & \multirow{2}{*}{\textbf{6.12}} \\
    & & \textcolor{tablegreen}{(8\%$\uparrow$)} & \textcolor{tablegreen}{(8\%$\uparrow$)} & \textcolor{tablegreen}{(11\%$\uparrow$)} & \textcolor{tablegreen}{(11\%$\uparrow$)} & \textcolor{tablegreen}{(10\%$\uparrow$)} & \\
    \midrule
    \multicolumn{8}{l}{\textbf{Qwen3-8B (Enable Thinking)}} \\
    \midrule
    \multirow{3}{*}{AIME25}
    & DFlash & 662 & 2121 & 3612 & 4956 & 5985 & 5.67 \\
    \cmidrule(lr){2-8}
    & \multirow{2}{*}{Draft-OPD}
    & \textbf{741} & \textbf{2465} & \textbf{4127} & \textbf{5729} & \textbf{6645}
    & \multirow{2}{*}{\textbf{6.42}} \\
    & & \textcolor{tablegreen}{(12\%$\uparrow$)} & \textcolor{tablegreen}{(16\%$\uparrow$)} & \textcolor{tablegreen}{(14\%$\uparrow$)} & \textcolor{tablegreen}{(16\%$\uparrow$)} & \textcolor{tablegreen}{(11\%$\uparrow$)} & \\
    \midrule
    \multirow{3}{*}{MATH-500}
    & DFlash & 703 & 2374 & 4154 & 5958 & 6991 & 5.99 \\
    \cmidrule(lr){2-8}
    & \multirow{2}{*}{Draft-OPD}
    & \textbf{787} & \textbf{2721} & \textbf{4691} & \textbf{6666} & \textbf{7940}
    & \multirow{2}{*}{\textbf{6.64}} \\
    & & \textcolor{tablegreen}{(12\%$\uparrow$)} & \textcolor{tablegreen}{(14\%$\uparrow$)} & \textcolor{tablegreen}{(13\%$\uparrow$)} & \textcolor{tablegreen}{(12\%$\uparrow$)} & \textcolor{tablegreen}{(13\%$\uparrow$)} & \\
    \midrule
    \multirow{3}{*}{SWE-Lite}
    & DFlash & 592 & 1962 & 3347 & 5051 & 6113 & 4.60 \\
    \cmidrule(lr){2-8}
    & \multirow{2}{*}{Draft-OPD}
    & \textbf{644} & \textbf{2263} & \textbf{3879} & \textbf{5611} & \textbf{6904}
    & \multirow{2}{*}{\textbf{5.27}} \\
    & & \textcolor{tablegreen}{(9\%$\uparrow$)} & \textcolor{tablegreen}{(15\%$\uparrow$)} & \textcolor{tablegreen}{(15\%$\uparrow$)} & \textcolor{tablegreen}{(11\%$\uparrow$)} & \textcolor{tablegreen}{(13\%$\uparrow$)} & \\
    \midrule
    \multicolumn{8}{l}{\textbf{Qwen3-30B-A3B-Thinking-2507}} \\
    \midrule
    \multirow{3}{*}{AIME25}
    & DFlash & 421 & 1086 & 1858 & 2738 & 4014 & 4.54 \\
    \cmidrule(lr){2-8}
    & \multirow{2}{*}{Draft-OPD}
    & \textbf{476} & \textbf{1229} & \textbf{2111} & \textbf{3187} & \textbf{4718}
    & \multirow{2}{*}{\textbf{5.32}} \\
    & & \textcolor{tablegreen}{(13\%$\uparrow$)} & \textcolor{tablegreen}{(13\%$\uparrow$)} & \textcolor{tablegreen}{(13\%$\uparrow$)} & \textcolor{tablegreen}{(16\%$\uparrow$)} & \textcolor{tablegreen}{(17\%$\uparrow$)} & \\
    \midrule
    \multirow{3}{*}{MATH-500}
    & DFlash & 417 & 1176 & 2009 & 3020 & 4462 & 5.44 \\
    \cmidrule(lr){2-8}
    & \multirow{2}{*}{Draft-OPD}
    & \textbf{477} & \textbf{1303} & \textbf{2243} & \textbf{3405} & \textbf{5007}
    & \multirow{2}{*}{\textbf{5.95}} \\
    & & \textcolor{tablegreen}{(14\%$\uparrow$)} & \textcolor{tablegreen}{(11\%$\uparrow$)} & \textcolor{tablegreen}{(11\%$\uparrow$)} & \textcolor{tablegreen}{(12\%$\uparrow$)} & \textcolor{tablegreen}{(12\%$\uparrow$)} & \\
    \midrule
    \multirow{3}{*}{SWE-Lite}
    & DFlash & 319 & 855 & 1484 & 2337 & 3453 & 3.43 \\
    \cmidrule(lr){2-8}
    & \multirow{2}{*}{Draft-OPD}
    & \textbf{352} & \textbf{960} & \textbf{1628} & \textbf{2579} & \textbf{3850}
    & \multirow{2}{*}{\textbf{3.81}} \\
    & & \textcolor{tablegreen}{(10\%$\uparrow$)} & \textcolor{tablegreen}{(11\%$\uparrow$)} & \textcolor{tablegreen}{(9\%$\uparrow$)} & \textcolor{tablegreen}{(10\%$\uparrow$)} & \textcolor{tablegreen}{(11\%$\uparrow$)} & \\
    \bottomrule
  \end{tabular}%
  }
  \caption{Throughput (tok/s), speedup over DFlash, and average acceptance length ($\tau$) on SGLang.}
  \label{tab:sglang-results}
\end{wraptable}

We benchmark Draft-OPD and DFlash on SGLang~\citep{zheng2024sglang} with the FA3 backend to evaluate deployment-time efficiency. Table~\ref{tab:sglang-results} reports serving throughput under concurrency levels from 1 to 32, together with the average acceptance length $\tau$.

Draft-OPD consistently improves SGLang serving efficiency across all evaluated models, tasks, and concurrency levels. It improves the acceptance length by 11.2\% on average over the evaluated model-task pairs and achieves up to a 17\% speedup on Qwen3-30B-A3B-Thinking. Notably, the throughput gains do not diminish under higher concurrency: at concurrency 32, the average relative gain is even higher than at concurrency 1. These serving results show that the higher acceptance lengths produced by Draft-OPD translate into practical throughput gains in an optimized inference engine.

\subsection{Ablation Study}
\label{sec:ablation-study}

\begin{wrapfigure}[10]{r}{0.5\textwidth}
  \centering
  \includegraphics[width=\linewidth]{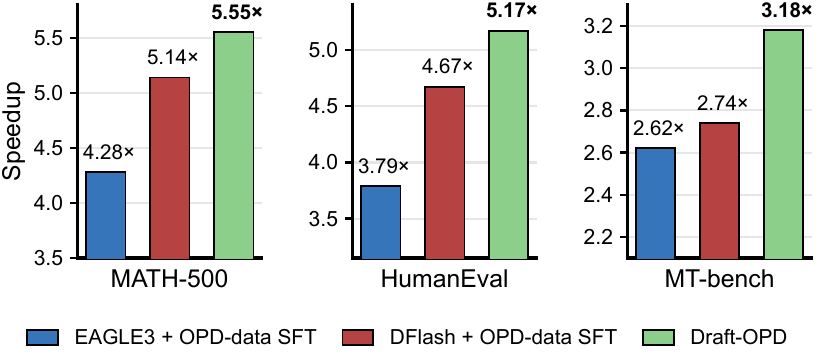}
  \caption{Training-data ablation on Qwen3-4B with thinking mode enabled.}
  \label{fig:training-data-ablation}
\end{wrapfigure}

\paragraph{Training Data.}
A key question is whether the gains of Draft-OPD come from on-policy distillation or simply from exposing the draft model to more prompts. Figure~\ref{fig:training-data-ablation} compares Draft-OPD with EAGLE-3 and DFlash variants that continue supervised training on responses generated by the target model from the OPD prompt pool. Draft-OPD performs better, showing that its improvement is not explained by additional supervised data alone. Instead, the benefit comes from distilling target distributions on states induced by the draft model during speculative decoding.

\paragraph{KL Type.}
The acceptance-aware KL objective is designed to match the asymmetric roles of accepted and rejected draft tokens. Table~\ref{tab:component-ablation} studies this design by replacing the mixed objective with all-forward and all-reverse KL variants. The all-forward variant is consistently below Draft-OPD, while the all-reverse variant performs the worst among the KL variants. This supports using different KL directions for accepted and rejected positions rather than applying a single divergence to all draft tokens.

\begin{wraptable}[9]{r}{0.5\textwidth}
  \centering
  \small
  \definecolor{draftopdblue}{RGB}{214,246,248}
  \setlength{\tabcolsep}{4pt}
  \resizebox{\linewidth}{!}{%
  \begin{tabular}{l*{6}{c}}
    \toprule
    \multirow{2}{*}{\raisebox{-0.45ex}{\textbf{Method}}}
    & \multicolumn{2}{c}{\textbf{MATH-500}}
    & \multicolumn{2}{c}{\textbf{HumanEval}}
    & \multicolumn{2}{c}{\textbf{MT-Bench}} \\
    \cmidrule(lr){2-3}
    \cmidrule(lr){4-5}
    \cmidrule(lr){6-7}
    & \textbf{Speedup} & \(\boldsymbol{\tau}\)
    & \textbf{Speedup} & \(\boldsymbol{\tau}\)
    & \textbf{Speedup} & \(\boldsymbol{\tau}\) \\
    \midrule
    \rowcolor{draftopdblue}
    Draft-OPD
    & \textbf{5.55$\times$} & \textbf{6.57}
    & \textbf{5.17$\times$} & \textbf{6.18}
    & \textbf{3.18$\times$} & \textbf{4.44} \\
    w/o Weight Decay
    & 5.13$\times$ & 6.18 & 4.96$\times$ & 6.01 & 3.07$\times$ & 4.29 \\
    All-reverse KL
    & 5.11$\times$ & 6.14 & 4.94$\times$ & 5.98 & 3.08$\times$ & 4.33 \\
    All-forward KL
    & 5.34$\times$ & 6.35 & 5.01$\times$ & 6.09 & 3.09$\times$ & 4.33 \\
    Random Anchors
    & 5.04$\times$ & 6.08 & 4.99$\times$ & 6.07 & 2.96$\times$ & 4.24 \\
    \bottomrule
  \end{tabular}%
  }
  \caption{Component ablation on Qwen3-4B.}
  \label{tab:component-ablation}
\end{wraptable}

\paragraph{Anchor Position.}
Error-position replay is intended to focus training on the states where speculative decoding actually fails. Table~\ref{tab:component-ablation} compares this design with randomly selected replay anchors. Random anchors lead to lower speedup and acceptance length, indicating that not all on-policy states are equally useful for draft-model post-training. Concentrating replay around rejected proposals better exposes the errors that determine speculative acceptance.

\paragraph{Weight Decay.}
The rejected-token position decay down-weights later rejected tokens in the same drafted block, since they are more likely to be affected by earlier draft errors. As shown in Table~\ref{tab:component-ablation}, removing this decay reduces both speedup and acceptance length. This suggests that the rejected-token loss is most effective when it emphasizes the earliest and most informative failure positions. We further analyze the robustness of Draft-OPD to the decay factor \(\gamma\) and relative rejected-token weight \(\lambda_{\mathrm{rej}}\) in Appendix~\ref{sec:hyperparameter-sensitivity}.

\subsection{Analysis Experiments}
\label{sec:target-assisted-rollout-analysis}

\paragraph{Target-Assisted Rollout.}
The direct draft-only rollout in Figure~\ref{fig:draft-opd-failure}(a) quickly becomes repetitive and low quality, making it unsuitable for OPD training despite being on-policy. We therefore analyze the naive target-assisted rollout baseline illustrated in Figure~\ref{fig:draft-opd-failure}(b). 

\begin{wraptable}[10]{r}{0.5\textwidth}
  \centering
  \small
  \definecolor{draftopdblue}{RGB}{214,246,248}
  \setlength{\tabcolsep}{4pt}
  \resizebox{\linewidth}{!}{%
  \begin{tabular}{l*{6}{c}}
    \toprule
    \multirow{2}{*}{\raisebox{-0.45ex}{\textbf{Setting}}}
    & \multicolumn{2}{c}{\textbf{MATH-500}}
    & \multicolumn{2}{c}{\textbf{HumanEval}}
    & \multicolumn{2}{c}{\textbf{MT-Bench}} \\
    \cmidrule(lr){2-3}
    \cmidrule(lr){4-5}
    \cmidrule(lr){6-7}
    & \textbf{Speedup} & \(\boldsymbol{\tau}\)
    & \textbf{Speedup} & \(\boldsymbol{\tau}\)
    & \textbf{Speedup} & \(\boldsymbol{\tau}\) \\
    \midrule
    \rowcolor{draftopdblue}
    Draft-OPD
    & \textbf{5.55$\times$} & \textbf{6.57}
    & \textbf{5.17$\times$} & \textbf{6.18}
    & \textbf{3.18$\times$} & \textbf{4.44} \\
    Naive rollout
    & 5.06$\times$ & 6.07
    & 4.80$\times$ & 5.93
    & 3.02$\times$ & 4.25 \\
    \bottomrule
  \end{tabular}%
  }
  \caption{Analysis of naive target-assisted rollout on Qwen3-4B.}
  \label{tab:target-assisted-rollout}
\end{wraptable}

Since speculative verification keeps the target-verified continuation and discards rejected draft proposals, this baseline reduces to KL-loss SFT on target-distributed trajectories. As shown in Table~\ref{tab:target-assisted-rollout}, naive target-assisted rollout underperforms Draft-OPD, reducing the average speedup from 4.63$\times$ to 4.29$\times$, a relative drop of 7.3\%. This confirms that stable target-assisted rollouts alone are insufficient; preserving draft-induced errors is important for effective online draft-model training.

\paragraph{Thinking-Mode Draftability Gap.}
We also observe that in Table~\ref{tab:main-results}, on math and code tasks, draft models trained under thinking mode are substantially less effective than their non-thinking counterparts. We attribute this gap to the higher uncertainty of target-model responses in thinking mode, where long reasoning traces often allow multiple plausible next steps, making it more difficult for the draft model to fit the target model's distribution. Appendix~\ref{sec:thinking-mode-gap} provides a slightly more detailed analysis.

\section{Discussion}
\label{sec:discussion}

We discuss two alternatives to the acceptance-aware objective: restricting supervision to the first rejected token and extending positional decay to accepted tokens.

\subsection{Training Only on the First Rejected Token}
\label{sec:first-rejected-only}

Draft-OPD applies the rejected-token objective to the suffix beginning at the first rejected token. Positions after the first rejection can be less reliable because they are conditioned on a draft prefix that has already been invalidated by the verifier. To assess whether these later positions provide useful supervision, we construct a \emph{First-rejected-only} variant that keeps the accepted-token objective unchanged but applies the rejected-token loss only to the first rejected token in each failed block.

\begin{table}[!ht]
  \centering
  \small
  \setlength{\tabcolsep}{6pt}
  \begin{tabular}{lcc}
    \toprule
    Benchmark & First-rejected-only & Draft-OPD \\
    \midrule
    GSM8K    & 6.32 & \textbf{6.39} \\
    AIME25   & \textbf{6.36} & 6.32 \\
    MATH-500 & \textbf{6.57} & \textbf{6.57} \\
    MBPP     & \textbf{5.81} & 5.73 \\
    HumanEval & \textbf{6.14} & 6.12 \\
    MT-Bench & 4.32 & \textbf{4.44} \\
    SWE-Lite & \textbf{6.08} & 6.07 \\
    \midrule
    Average  & 5.94 & \textbf{5.95} \\
    \bottomrule
  \end{tabular}
  \caption{Average acceptance length \(\tau\) when applying the rejected-token loss only to the first rejected token.}
  \label{tab:first-rejected-only}
\end{table}

As shown in Table~\ref{tab:first-rejected-only}, First-rejected-only and Draft-OPD achieve nearly identical average acceptance lengths of 5.94 and 5.95, respectively. This result suggests that the first rejected token captures most of the useful error signal, whereas later suffix tokens provide limited additional benefit and may introduce noise. Moreover, this objective is compatible with a truncated replay implementation that stops after scoring the first failure, potentially reducing replay-stage computation. First-rejected-only is therefore a promising efficiency-oriented simplification of Draft-OPD.

\subsection{Why Decay Is Applied Only to Rejected Tokens}
\label{sec:accepted-token-decay}

The positional decay in Draft-OPD is intended to account for the decreasing reliability of rejected suffixes, rather than to impose a general preference for earlier block positions. To test whether accepted tokens also benefit from positional decay, we construct an \emph{Accept-decay} variant that applies the same decay factor \(\gamma=0.8\) to the accepted-token objective while keeping all other settings unchanged.

\begin{table}[!ht]
  \centering
  \small
  \setlength{\tabcolsep}{6pt}
  \begin{tabular}{lcc}
    \toprule
    Benchmark & Accept-decay & Draft-OPD \\
    \midrule
    GSM8K    & 6.25 & \textbf{6.39} \\
    AIME25   & 6.28 & \textbf{6.32} \\
    MATH-500 & 6.44 & \textbf{6.57} \\
    MBPP     & 5.66 & \textbf{5.73} \\
    HumanEval & 6.11 & \textbf{6.12} \\
    MT-Bench & 4.33 & \textbf{4.44} \\
    SWE-Lite & 5.97 & \textbf{6.07} \\
    \midrule
    Average  & 5.86 & \textbf{5.95} \\
    \bottomrule
  \end{tabular}
  \caption{Average acceptance length \(\tau\) when positional decay is also applied to accepted tokens.}
  \label{tab:accepted-token-decay}
\end{table}

Table~\ref{tab:accepted-token-decay} shows that Accept-decay reduces the acceptance length on all seven benchmarks, lowering the average from 5.95 to 5.86. Accepted tokens remain on the verified target trajectory and therefore provide reliable supervision throughout a block. Applying decay to positions later within the accepted portion of the block unnecessarily down-weights this reliable supervision, explaining why decay is beneficial for rejected suffixes but not for accepted tokens.

\section{Conclusion}
\label{sec:conclusion}

We propose Draft-OPD, an on-policy distillation framework for training-based draft models. By using target-assisted rollouts with an error-position replay mechanism, Draft-OPD keeps training samples stable while preserving the draft-policy errors that determine speculative acceptance. We further introduce an acceptance-aware distillation objective that treats accepted and rejected draft tokens differently, enabling the draft model to learn from both reliable proposals and informative failure modes. Experiments on Qwen3 models show that Draft-OPD improves average acceptance length and end-to-end decoding speedup over strong draft-model baselines under a matched training budget, with additional SGLang results confirming practical serving gains. Overall, Draft-OPD shows that draft models benefit from post-training on verification-time errors, offering an effective direction for improving training-based draft model.

\section*{Limitations}

\paragraph{Training Length.}
Due to compute constraints, for thinking-mode Draft-OPD training we cap the maximum response length at 4096 tokens, while evaluation uses 8192 tokens. Although Draft-OPD still achieves strong gains under this longer evaluation setting, the training rollouts may not fully cover late-stage states in very long generations. Scaling OPD training to longer rollouts could expose the draft model to a broader range of verification-time errors and may further improve draft-target alignment for long outputs.

\paragraph{Evaluation Scope.}
This work focuses on post-training draft models for lossless speculative decoding. Our main experiments are conducted on Qwen3 models, and the Draft-OPD implementation follows the DFlash-style parallel draft architecture. Although we include comparisons with EAGLE-3 and evaluate deployment in SGLang, further evaluation is needed to understand how well Draft-OPD transfers to other model families, draft architectures, and inference backends.

\paragraph{Lossless Decoding.}
Because speculative decoding preserves the target model distribution, Draft-OPD is designed to improve decoding efficiency rather than generation quality; extending on-policy draft-model training to settings with approximate or lossy verification is an interesting direction for future work.

\section*{Acknowledgments}

This work was supported by the Shanghai Artificial Intelligence Laboratory. We are grateful to the authors and open-source communities whose work made this project possible. In particular, DFlash~\citep{chen2026dflash} and EAGLE-3~\citep{li2025eagle3} provided strong foundations and reference points for training-based speculative decoding, while SGLang~\citep{zheng2024sglang}, SpecForge~\citep{li2026specforge}, and verl~\citep{sheng2024hybridflow} offered practical infrastructure for serving and training experiments. We especially appreciate SpecForge for its open-source implementation of DFlash training, which provided a useful reference for our training setup. We also thank Runzhe Zhan for sharing valuable experience on OPD training and for discussions that helped us stabilize the training workflow.

\bibliography{paper}
\bibliographystyle{iclr2026_conference}
\appendix
\section{Training Details}
\label{sec:opd-training-details}

\paragraph{SFT stage.}
We follow the SFT configuration of DFlash \citep{chen2026dflash} for most training settings and use SpecForge \citep{li2026specforge} as the training framework. For Draft-OPD, we initialize the OPD stage from the draft-model checkpoint after 6 SFT epochs. For the EAGLE-3 and DFlash baselines used in our comparisons, we continue supervised draft-model training for 10 epochs under the same data setup and report the checkpoint with the best evaluation performance.

\paragraph{OPD stage.}
For Draft-OPD training, we use the rejected-token position weights
\[
  w_k = \gamma^{k-1},
\]
with \(\gamma=0.8\). We train on the OPD data mixture described in Section~\ref{sec:experiment} for 8 epochs, using a maximum response length of 4096 tokens for thinking-enabled and 2048 tokens for thinking-disabled. Optimization uses AdamW with a learning rate of \(3\times 10^{-4}\) and a cosine learning-rate schedule with a warmup ratio of 0.05. We implement the OPD stage with verl \citep{sheng2024hybridflow}.
For the final Draft-OPD objective in Equation~\ref{eq:draft-opd-objective}, we set \(\lambda_{\mathrm{acc}}=\lambda_{\mathrm{rej}}=1\) in all experiments.

\paragraph{Training cost and reproducibility.}
All Draft-OPD experiments use NVIDIA H200 GPUs. The Qwen3-4B and Qwen3-8B experiments use 8 GPUs, while Qwen3-30B-A3B-Thinking uses 16 GPUs. For Qwen3-4B, the OPD stage takes approximately 15 hours on 8 H200 GPUs, corresponding to approximately 120 H200 GPU-hours. This accounting includes the frozen target model's rollout and verification computation. Table~\ref{tab:opd-training-cost} summarizes the Qwen3-4B OPD configuration.

\begin{table}[t]
  \centering
  \small
  \setlength{\tabcolsep}{6pt}
  \caption{Training cost and configuration of the Qwen3-4B OPD stage.}
  \label{tab:opd-training-cost}
  \begin{tabular}{ll}
    \toprule
    Item & Qwen3-4B OPD setting \\
    \midrule
    GPU hardware & \(8\times\) NVIDIA H200 \\
    OPD wall-clock time & \(\sim\)15 hours \\
    OPD GPU-hours & \(\sim\)120 H200 GPU-hours \\
    Prompt pool & 16K prompts, disjoint from SFT data \\
    Global training batch size & 24 \\
    OPD steps & \(\sim\)5.3K \\
    Epochs / prompt reuse & 8 epochs / 8 passes per prompt \\
    Maximum rollout length & 4096 (thinking), 2048 (non-thinking) \\
    \bottomrule
  \end{tabular}
\end{table}

\section{Loss Design for Accepted and Rejected Draft Tokens}
\label{sec:loss-design}

We provide a local training-objective justification for using different KL directions on accepted and rejected replay positions. The goal is not to show that the mixed KL objective universally dominates all-forward or all-reverse KL. For a fixed replay state and an unconstrained draft distribution, these objectives share the same optimum \(q=p\). Their difference lies in how the local loss weights target-supported tokens and draft-proposed tokens during finite-capacity optimization.

Fix a replay state \(s\) and let \(p(\cdot\mid s)\) and \(q(\cdot\mid s)\) denote the target and draft next-token distributions over vocabulary \(\mathcal{V}\). In the following derivation, all probabilities are conditioned on \(s\), so we write \(p(y)\) and \(q(y)\) for brevity. For an accepted replay position, the draft proposal has passed target verification, and the local objective should match the target distribution at this reliable state. This gives the target-weighted cross-entropy
\begin{equation}
  \mathcal{J}_{\mathrm{acc}}(q)
  =
  \mathbb{E}_{y\sim p}[-\log q(y)] .
\end{equation}
Expanding the expectation yields
\begin{align}
  \mathcal{J}_{\mathrm{acc}}(q)
  &=
  -\sum_{y\in\mathcal{V}} p(y)\log q(y) \\
  &=
  H(p) + D_{\mathrm{KL}}(p\,\|\,q),
\end{align}
where \(H(p)\) is independent of \(q\). Thus, the accepted-position objective is equivalent, up to an additive constant, to minimizing forward KL.

For a rejected replay position, the relevant signal is draft-induced: the token belongs to a draft-proposed suffix that failed or was invalidated by target verification. The loss should therefore place weight on tokens that the draft model itself is likely to propose, especially when those modes are not supported by the target. This leads to the draft-weighted disagreement objective
\begin{equation}
  \mathcal{J}_{\mathrm{rej}}(q)
  =
  \mathbb{E}_{y\sim q}\!\left[\log \frac{q(y)}{p(y)}\right].
\end{equation}
Since
\begin{equation}
  \mathcal{J}_{\mathrm{rej}}(q)
  =
  \sum_{y\in\mathcal{V}} q(y)\log \frac{q(y)}{p(y)}
  =
  D_{\mathrm{KL}}(q\,\|\,p),
\end{equation}
the rejected-position objective corresponds to reverse KL. Unlike the accepted-position objective, this loss is weighted by the draft distribution and therefore directly penalizes high-probability draft modes that disagree with the target distribution.

Aggregating these local objectives over replay positions gives the acceptance-aware form used in Draft-OPD:
\begin{equation}
  \sum_{i\in\mathcal{I}_{\mathrm{acc}}}
  D_{\mathrm{KL}}(p_i\,\|\,q_i)
  +
  \sum_{i\in\mathcal{I}_{\mathrm{rej}}}
  w_i D_{\mathrm{KL}}(q_i\,\|\,p_i),
\end{equation}
up to normalization and scalar weights. This decomposition clarifies the role of the two KL directions. Forward KL is appropriate for accepted positions because the supervision is target-weighted on verified states, while reverse KL is appropriate for rejected positions because the supervision is draft-weighted on states that expose draft errors. Applying a single KL direction to all positions ignores this distinction: all-forward KL treats rejected draft errors like reliable accepted states, whereas all-reverse KL treats verified accepted positions like draft-error states. The component ablations in Table~\ref{tab:component-ablation} are consistent with this local objective view.

\section{Hyperparameter Sensitivity}
\label{sec:hyperparameter-sensitivity}

We study the sensitivity of Draft-OPD to the rejected-token decay factor \(\gamma\) and the relative rejected-token weight \(\lambda_{\mathrm{rej}}\). In each sweep, we vary one hyperparameter while keeping all other settings fixed and report the average acceptance length \(\tau\) on four benchmarks.

\paragraph{Decay factor \(\gamma\).}
We sweep \(\gamma\) from 0.5 to 0.95. As shown in Table~\ref{tab:gamma-sensitivity}, the average acceptance length remains stable between 6.24 and 6.26 across the entire sweep. In particular, values between 0.5 and 0.8 achieve comparable performance, indicating that Draft-OPD is not sensitive to the precise decay factor.

\begin{table}[!ht]
  \centering
  \small
  \setlength{\tabcolsep}{5pt}
  \caption{Sensitivity to the rejected-token decay factor \(\gamma\). All entries report average acceptance length \(\tau\).}
  \label{tab:gamma-sensitivity}
  \begin{tabular}{lcccc}
    \toprule
    \(\gamma\) & 0.95 & 0.80 & 0.65 & 0.50 \\
    \midrule
    AIME25   & 6.30 & 6.32 & \textbf{6.34} & 6.31 \\
    MBPP     & 5.69 & \textbf{5.73} & 5.71 & 5.70 \\
    MATH-500 & \textbf{6.58} & 6.57 & 6.56 & 6.54 \\
    GSM8K    & 6.37 & 6.39 & \textbf{6.41} & 6.39 \\
    \midrule
    Average  & 6.24 & 6.25 & \textbf{6.26} & 6.24 \\
    \bottomrule
  \end{tabular}
\end{table}

\paragraph{Relative rejected-token weight.}
We fix \(\lambda_{\mathrm{acc}}=1\) and vary \(\lambda_{\mathrm{rej}}\). Table~\ref{tab:rejected-weight-sensitivity} shows that the strongest region is \(\lambda_{\mathrm{rej}}\in[0.5,1]\), with \(\lambda_{\mathrm{rej}}=0.8\) attaining the highest average in this sweep. Larger weights that overemphasize rejected tokens are modestly worse. Overall, performance is stable when the rejected-token weight does not exceed the accepted-token weight, and the default \(\lambda_{\mathrm{rej}}=1\) used in our main experiments lies within this robust range.

\begin{table}[!ht]
  \centering
  \small
  \setlength{\tabcolsep}{4pt}
  \caption{Sensitivity to the relative rejected-token weight with \(\lambda_{\mathrm{acc}}=1\). All entries report average acceptance length \(\tau\).}
  \label{tab:rejected-weight-sensitivity}
  \begin{tabular}{lccccc}
    \toprule
    \(\lambda_{\mathrm{rej}}\) & 2.0 & 1.5 & 1.0 & 0.8 & 0.5 \\
    \midrule
    AIME25   & 6.25 & 6.27 & 6.32 & \textbf{6.35} & 6.30 \\
    MBPP     & 5.69 & 5.72 & 5.73 & \textbf{5.79} & 5.71 \\
    MATH-500 & 6.51 & 6.53 & \textbf{6.57} & 6.55 & 6.52 \\
    GSM8K    & 6.34 & 6.38 & 6.39 & 6.37 & \textbf{6.41} \\
    \midrule
    Average  & 6.20 & 6.23 & 6.25 & \textbf{6.27} & 6.24 \\
    \bottomrule
  \end{tabular}
\end{table}

\section{Replay Distribution Analysis}
\label{sec:replay-distribution}

We analyze how the verification outcomes associated with replay anchors evolve during OPD training. For each replayed draft block of size \(K=16\), we record the position of the first rejected token within the block. The \emph{Full accept} category denotes that all 16 draft tokens are accepted. Table~\ref{tab:replay-position-distribution} compares the resulting distribution over the first and last halves of OPD training.

\begin{table}[!ht]
  \centering
  \small
  \setlength{\tabcolsep}{5pt}
  \caption{Distribution of the first-rejection position within replayed draft blocks during the first and last halves of OPD training.}
  \label{tab:replay-position-distribution}
  \begin{tabular}{lcc@{\hspace{2em}}lcc}
    \toprule
    Position & First 50\% & Last 50\% & Position & First 50\% & Last 50\% \\
    \midrule
    1 & 12.61\% & 11.98\% & 9  & 4.03\% & 4.18\% \\
    2 & 14.34\% & 13.86\% & 10 & 3.37\% & 3.50\% \\
    3 & 12.65\% & 12.38\% & 11 & 2.81\% & 2.92\% \\
    4 & 10.53\% & 10.44\% & 12 & 2.37\% & 2.47\% \\
    5 &  8.61\% &  8.63\% & 13 & 1.99\% & 2.09\% \\
    6 &  7.09\% &  7.19\% & 14 & 1.70\% & 1.78\% \\
    7 &  5.85\% &  5.99\% & 15 & 1.49\% & 1.55\% \\
    8 &  4.87\% &  5.03\% & Full accept & 5.68\% & 6.02\% \\
    \bottomrule
  \end{tabular}
\end{table}

The first-rejection distribution shifts toward later positions during OPD. The probability mass at positions 1--3 decreases from 39.61\% to 38.23\%, whereas the mass at positions 7--15 increases from 28.48\% to 29.50\%. The full-accept rate also rises from 5.68\% to 6.02\%. Since a later first rejection corresponds to more accepted draft tokens within a block, this rightward shift reflects improved draft--target alignment and acceptance over the course of OPD training.

\newpage
\section{Thinking-Mode Drafting Gap}
\label{sec:thinking-mode-gap}

\begin{wrapfigure}{r}{0.5\textwidth}
  \vspace{-3.0em}
  \centering
  \includegraphics[width=\linewidth]{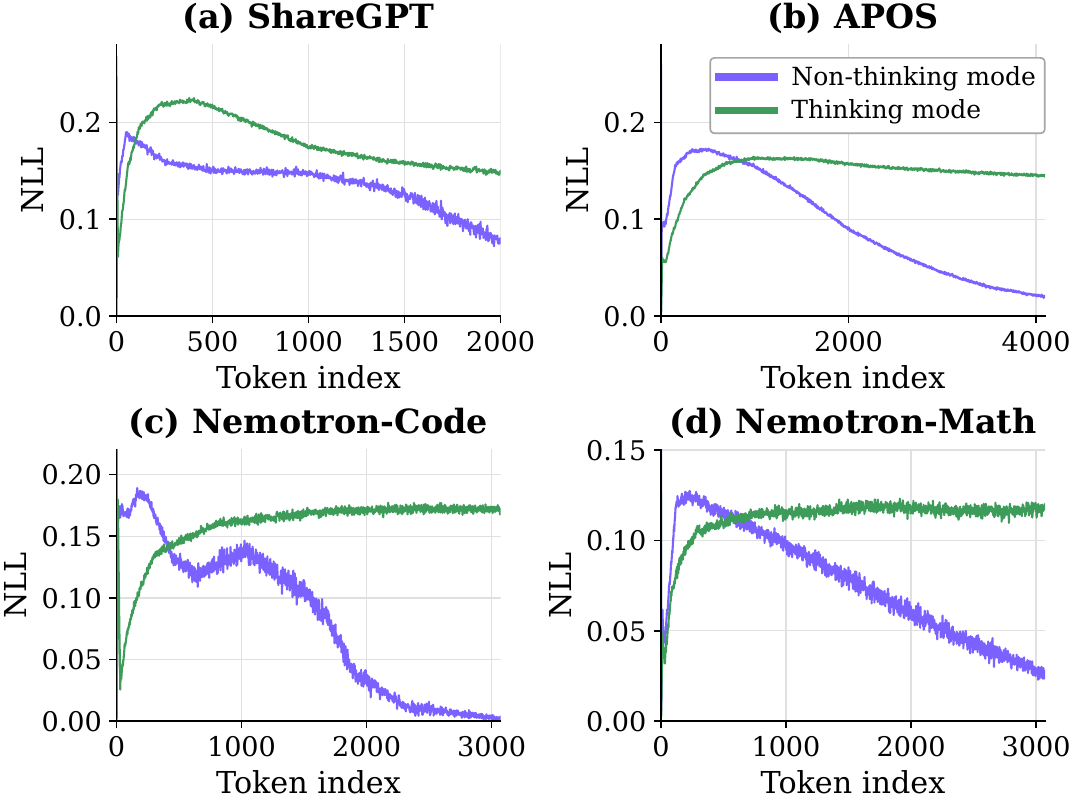}
  \caption{Token-level negative log-likelihood under thinking and non-thinking modes across evaluated datasets. }
  \label{fig:thinking-mode-nll}
\end{wrapfigure}

To examine this gap, we run Qwen3-4B in both thinking and non-thinking modes on prompts from ShareGPT \citep{sharegpt}, AoPS \citep{aops}, the math and code splits of Nemotron-Post-Training-Dataset-v2 \citep{nemotronposttrainingdatasetv2}, and compute the next-token NLL at each token position for the generated responses. Figure~\ref{fig:thinking-mode-nll} shows that thinking-mode responses have higher next-token NLL than non-thinking responses across the evaluated datasets. This suggests that, in thinking mode, the target model itself has higher uncertainty over the next token. A draft model trained for thinking-mode decoding is therefore asked to predict a less concentrated target distribution, making it harder to match the target model's subsequent tokens accurately.

Draft-OPD partially mitigates this challenge by post-training the draft model on verification-time errors, improving acceptance for reasoning-oriented decoding; however, since our method is designed as a general post-training framework across decoding settings, approaches tailored specifically to reasoning draft models remain an important direction for future work.

\end{document}